\documentclass[letterpaper]{article} 
\usepackage{aaai25}  
\usepackage{times}  
\usepackage{helvet}  
\usepackage{courier}  
\usepackage[hyphens]{url}  
\usepackage{graphicx} 
\usepackage{natbib}  
\usepackage{caption} 
\usepackage{algorithm}
\usepackage{algorithmic}

\usepackage{newfloat}
\usepackage{listings}
\usepackage{threeparttable}
\usepackage{multirow} 
\usepackage{adjustbox}
\usepackage{amsmath}
\usepackage{amssymb}

\DeclareCaptionStyle{ruled}{labelfont=normalfont,labelsep=colon,strut=off} 
\floatstyle{ruled}
\newfloat{listing}{tb}{lst}{}
\floatname{listing}{Listing}
\nocopyright

\title{SpikingNeRF: Making Bio-inspired Neural Networks See through the Real World}

\author{
Xingting Yao$^{1,2}$\thanks{Equal contribution.},  
Qinghao Hu$^{1*}$,
Fei Zhou$^{3}$,
Tielong Liu$^{1,2}$, \\
Zitao Mo$^{1}$, 
Zeyu Zhu$^{1,2}$, 
Zhengyang Zhuge$^{1}$, 
Jian Cheng$^{1}$\thanks{Corresponding author.}\\
$^1$Institute of Automation, Chinese Academy of Sciences\\
$^2$School of Future Technology, University of Chinese Academy of Sciences\\
$^3$China Electric Power Research Institute Co., Ltd\\
{\tt\small  \{yaoxingting2020, huqinghao2014, jian.cheng\}@ia.ac.cn,}
}

\usepackage{bibentry}

\begin{document}

\maketitle

\begin{abstract}
In this paper, we propose SpikingNeRF, which aligns the temporal dimension of spiking neural networks (SNNs) with the radiance rays, to seamlessly accommodate SNNs to the reconstruction of neural radiance fields (NeRF). Thus, the computation turns into a spike-based, multiplication-free manner, reducing energy consumption and making high-quality 3D rendering, for the first time, accessible to neuromorphic hardware. In SpikingNeRF, each sampled point on the ray is matched to a particular time step and represented in a hybrid manner where the voxel grids are maintained as well.
Based on the voxel grids, sampled points are determined whether to be masked out for faster training and inference. 
However, this masking operation also incurs irregular temporal length, making it intractable for hardware processors, e.g., GPUs, to conduct parallel training.
To address this problem, we develop the temporal padding strategy to tackle the masked samples to maintain regular temporal length, i.e., regular tensors, and further propose the temporal condensing strategy to form a denser data structure for hardware-friendly computation. 
Experiments on various datasets demonstrate that our method can reduce energy consumption by an average of 70.79\% and obtain comparable synthesis quality with the ANN baseline. Verification on the neuromorphic hardware accelerator also shows that SpikingNeRF can further benefit from neuromorphic computing over the ANN baselines on energy efficiency. Codes and the appendix are in \url{https://github.com/Ikarosy/SpikingNeRF-of-CASIA}.
\end{abstract}

\section{Introduction}
\label{sec:intro}

Spiking neural networks (SNNs) are considered the third generation of neural networks, and their bionic modeling encourages much research attention to explore the prospective bio-plausible intelligence that features multitasking and extreme energy efficiency as the human brain does \cite{maass1997networks,roy2019towards}.
While much dedication has been devoted to SNN research, the gap between the expectation of SNN boosting a wider range of intelligent tasks and the fact of artificial neural networks (ANNs) dominating deep learning in the majority of tasks still exists. 

Recently, more research interests have been invested in narrowing the gap and have promoted notable milestones in various tasks, including image classification \cite{zhou2022spikformer}, object detection \cite{zhang2022spiking}, graph prediction \cite{zhu2022spiking}, natural language processing \cite{zhu2023spikegpt}, etc. Besides  multi-task supporting, SNN 
research is also thriving
\begin{figure}[t]
  \centering
  \includegraphics[width= 8cm]{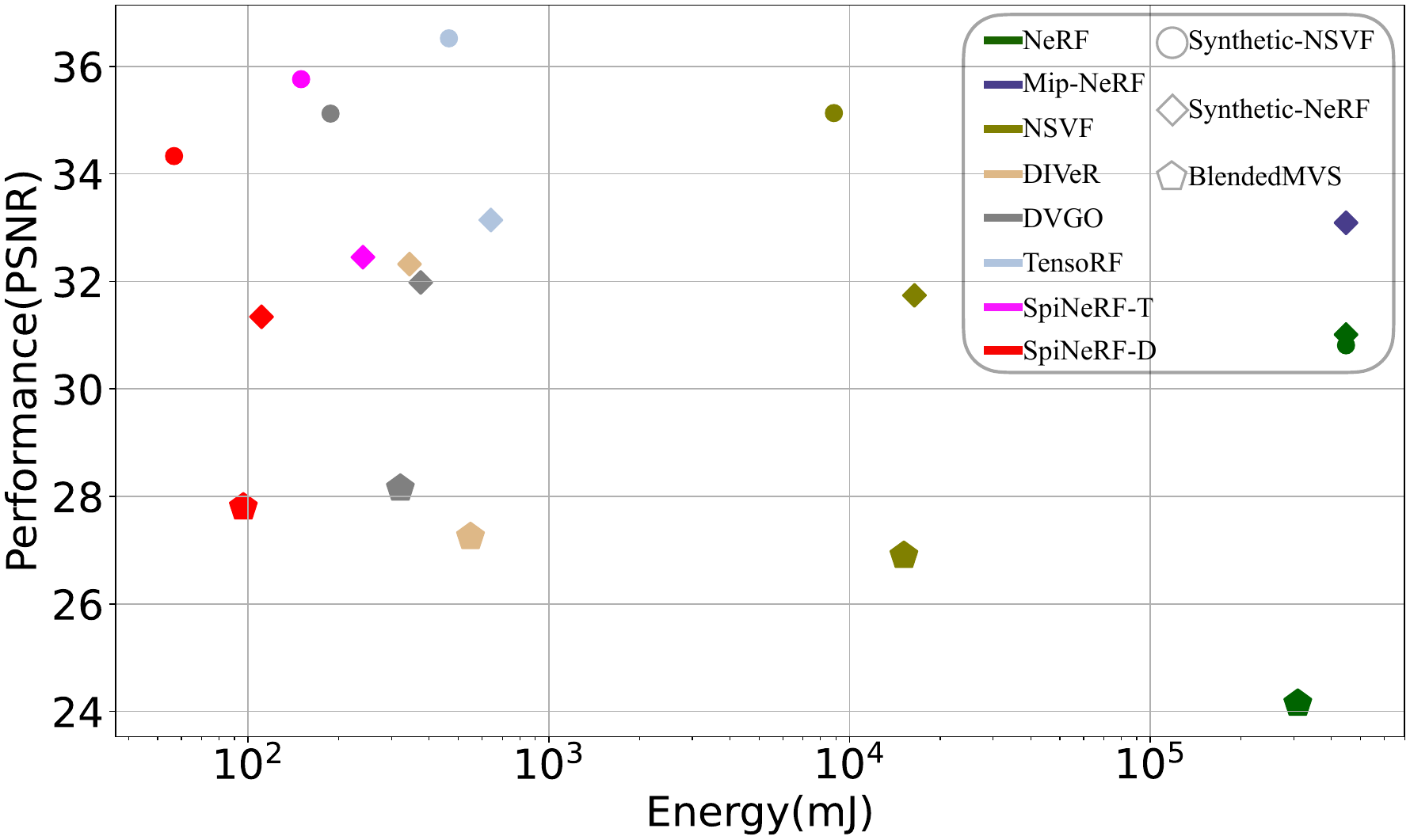}
  \caption{Comparisons of our SpikingNeRF with other NeRF-based works in synthesis quality and model rendering energy. Different colors represent different works, and our SpikingNeRF with two different frameworks are denoted in red and violet, respectively. A detailed notation explanation is specified in the Experiments section. Different testing datasets are denoted by different shapes.}
  \label{fig:performance-energy}
\end{figure} 
in performance lifting and energy efficiency exploration at the same time.

However, we have not yet witnessed the establishment of SNN in the real 3D reconstruction task with advanced 
performance. Let alone enabling the high-quality real 3D rendering on neuromorphic hardwares, e.g., Loihi \cite{davies2018loihi}, PTB \cite{lee2022parallel}, and SpikeSim \cite{10122627}, where neuromorphic computing can essentially acquire low-energy consumption. Meanwhile, high-quality real 3D reconstruction, specifically for NeRF \cite{mildenhall2021nerf}, has suffered huge computation overhead during rendering, consuming a significant magnitude of energy \cite{garbin2021fastnerf}. 
Naturally, this raises a question: \emph{could bio-inspired spiking neural networks reconstruct the real 3D scene with advanced quality at low energy consumption}? This paper investigates the rendering of neural radiance fields with a spiking approach to answer the question.

We propose SpikingNeRF to reconstruct volumetric scene representations of neural radiance fields. For fast synthesis, voxel grids methods \cite{hedman2021baking, liu2022neural, sun2022direct} are considered to explicitly store the volumetric parameters. For efficient computation, the spiking multilayer perceptron (sMLP) is utilized to implicitly yield volumetric information in an addition-only and spike-driven approach. With such an explicit-and-implicit hybrid, fast and energy-efficient neural radiance rendering becomes feasible. 

Inspired by the imaging process of the primate fovea in the retina that accumulates the intensity flow over time to stimulate the photoreceptor cell \cite{masland2012neuronal, wassle2004parallel}, we take one step forward to associate the accumulation process of rendering with the temporal accumulation process of SNNs, which ultimately stimulates the spiking neurons to fire. Concretely, we align the radiance ray with the temporal dimension of the sMLP, and individually match each sampled point on the ray to a time step during rendering.  Thus, the geometric consecutiveness of the ray is transformed into the temporal continuity of the SNN. As a result, SpikingNeRF seizes the nature of both worlds to make the NeRF rendering in a spiking manner, and brings about a novel and effective data encoding approach for SNN-based 3D rendering. Different from other SNN-based image reconstructions \cite{zhu2022event, mei2023deep}, which focus on the event-based gray 2D reconstruction, we are the first to explore the reconstruction of the real RGB world with SNNs.

Moreover, since the number of sampled points on different rays always varies, the temporal lengths of different rays become irregular. Consequently, the querying for the volumetric information can hardly be parallelized during rendering, severely hindering the training process  
on \emph{graphics processing units (GPUs)}. 
To solve this issue, we first investigate the temporal padding (TP) method to attain the regular temporal length in a querying batch, i.e., a regular-shaped tensor, thus ensuring parallelism and GPU training feasible. 
Furthermore, we propose the temporal condensing-and-padding (TCP), to fully constrain the tensor size and condense the data distribution, which is hardware-friendly to \emph{neuromorphic hardware accelerators} and \emph{GPUs}.
Our thorough experimentation proves that TCP can maintain both the energy merits of SNNs and the high quality of NeRF rendering as shown in Fig. \ref{fig:performance-energy}.

To sum up, our main contributions are as follows:
\begin{itemize}
    \item We propose SpikingNeRF that aligns the temporal dimension of SNNs with the radiance rays of NeRF, exploiting the temporal characteristics of SNNs. \emph{To the best of our knowledge, this is the first work to accommodate SNNs to reconstructing 3D scenes, making high-quality 3D rendering feasible on the neuromorphic hardware}.
    \item We propose TP and TCP to solve the irregular temporal lengths, ensuring the training and inference parallelism on GPUs. TCP can also further keep SpikingNeRF hardware-friendly to the neuromorphic hardware.
    \item Our experiments demonstrate the effectiveness of SpikingNeRF on four mainstream tasks, achieving advanced energy efficiency as shown in Fig. \ref{fig:performance-energy}. For another specific example, SpikingNeRF-D can achieve 72.95\% energy reduction with a 0.33 PSNR drop on Tanks\&Temples.
\end{itemize}

In order to avoid misunderstanding, we additionally cite  \cite{liao2023spiking} which shares the same title as ours. Their work, based on the ANN implementation, essentially builds a non-linear and non-spike function named B-FIF to post-process the particular density-related output of the original ANN-based NeRF. Overall,  they do not use SNNs, do not aim at the neuromorphic hardware, focus on geometric reconstruction, and report only on the Chamfer metric. So, it is rational to deem that \cite{liao2023spiking} is irrelevant and our work is impossible to conduct quantitative comparisons with theirs. Different from the above work, we are the first to explore the rendering of real RGB with SNNs and benefit NeRF rendering from neuromorphic computing.

\section{Preliminaries}
\label{sec:Preliminaries}

\textbf{Neural radiance field.}
To reconstruct the scene for the given view, NeRF \cite{mildenhall2021nerf} first utilizes an MLP, which takes in the location coordinates $\mathbf{p}\in\mathbb{R}^3$ and the view direction $\mathbf{v}\in\mathbb{R}^2$ and yields the density $\sigma\in\mathbb{R}$ and the color $\mathbf{c}\in\mathbb{R}^3$, to implicitly maintain continuous volumetric representations:
\begin{align}
  \mathbf{e}, \sigma &= MLP_{\theta }(\mathbf{p}),
  \label{eq:nerf mlp1}
  \\
  \mathbf{c} &= MLP_{\gamma }(\mathbf{e}, \mathbf{v}),
  \label{eq:nerf mlp2}
\end{align}
where $\theta$ and $\gamma$ denote the parameters of the separate two parts of the MLP, and $\mathbf{e}$ is the embedded features. Next, NeRF renders the pixel of the expected scene by casting a ray $\mathbf{r}$ from the camera origin point to the direction of the pixel, then sampling $K$ points along the ray. Through querying the MLP as in Eq. (\ref{eq:nerf mlp1}-\ref{eq:nerf mlp2}) $K$ times, $K$ color values and $K$ density values can be retrieved. Finally, following the principles of the discrete volume rendering proposed in  \cite{max1995optical}, the expected pixel RGB $\hat{C}(\mathbf{r})$ can be rendered:
\begin{align}
    \alpha = 1-\exp(-\sigma_i&\delta_i),\quad
    w_i=\prod^{i-1}_{j=1}(1-\alpha_i), \label{eq:nerf weights}\\
    \hat{C}(\mathbf{r})&\approx\sum^K_{i=1}w_i\alpha_i \mathbf{c}_i, \label{eq:nerf accumulation}
\end{align}
where $\mathbf{c}_i$ and $\sigma_i$ denotes the color and the density values of the $i$-th point respectively, and $\delta_i$ is the distance between the adjacent point $i$ and $i+1$.

After rendering all the pixels, the expected scene is reconstructed. With the ground-truth pixel color $C(\mathbf{r})$, the parameters of the MLP can be trained end-to-end by minimizing the MSE loss:
\begin{equation}
    \mathcal{L} = \frac{1}{|\mathcal{R}|}\sum_{r\in \mathcal{R}}\Vert\hat{C}(\mathbf{r})-C(\mathbf{r})\Vert^2_2,
    \label{eq:nerf mse}
\end{equation}
where $\mathcal{R}$ is the mini-batch containing the sampled rays.
\\\textbf{Hybrid volumetric representation.}
The number of sampled points $K$ in Eq. \ref{eq:nerf accumulation} is usually big, leading to the heavy MLP querying burden as displayed in Eq. (\ref{eq:nerf mlp1}-\ref{eq:nerf mlp2}). To alleviate this problem, voxel grid representation is utilized to contain the volumetric parameters directly, e.g., the embedded feature $e$ and the density $\sigma$ in Eq. \ref{eq:nerf mlp1}, as the values of the voxel grid. Thus, querying the MLP in Eq. \ref{eq:nerf mlp1} is substituted to querying the voxel grids and operating the interpolation, which is much easier:
\begin{align}
    &\sigma = \mathrm{act}(\mathrm{interp}(\mathbf{p},\mathbf{V_{\sigma}})),\label{eq:voxel optimization sigma}\\
    &\mathbf{e} = \mathrm{interp}(\mathbf{p},\mathbf{V_{f}}), \label{eq:voxel optimization embed}
\end{align}
where $\mathbf{V_{\sigma}}$ and $\mathbf{V_{f}}$ are the voxel grids related to the volumetric density and features, respectively. “interp” denotes the interpolation operation, and “act” refers to the activation function, e.g., ReLU or the shifted softplus \cite{barron2021mip}.

Furthermore, those irrelevant points with low density or unimportant points with low weight can be \textbf{masked} through predefined thresholds $\lambda$, then Eq. \ref{eq:nerf accumulation} turns into:
\begin{align}
&A \triangleq \{i: w_i > \lambda_1, \alpha_i> \lambda_2 \},\label{eq:voxel mask1}\\
&\hat{C}(\mathbf{r})\approx\sum_{i \in A} w_i\alpha_i \mathbf{c}_i.\label{eq:voxel mask2}
\end{align}
Thus, the queries of the MLP for sampled points in Eq. \ref{eq:nerf mlp2} are significantly reduced. With such computational benefits, hybrid volumetric representation is prevalent in neural radiance rendering  \cite{sun2022direct,chen2022tensorf}.
\\\textbf{Spiking neuron.}
The spiking neuron is the most fundamental unit in spiking neural networks, which essentially differs SNNs from ANNs. The modeling of spiking neurons commonly adopts the leaky integrate-and-fire (LIF) model:
\begin{align}
        \label{eq:lif 1}
        \mathbf{U}^{t} &= \mathbf{V}^{t-1} + \frac{1}{\tau}(\mathbf{X}^t-\mathbf{V}^{t-1}+V_{reset}) ,
        \\
        \label{eq:lif 2}
        \mathbf{S}^{t} &= \mathbb{H}( \mathbf{U}^{t} - V_{th}),\\
        \label{eq:lif 3}
        \mathbf{V}^{t} &= \mathbf{U}^t\odot(1-\mathbf{S}^{t}) + V_{reset} \mathbf{S}^{t}.
\end{align}
Here, we follow the renowned SpikingJelly \cite{fang2023spikingjelly} to implement the LIF neurons. $\odot$ denotes the Hadamard product.  $\mathbf{U}^{t}$ is the intermediate membrane potential at time-step $t$ and can be updated through Eq. \ref{eq:lif 1}, where $\mathbf{V}^{t-1}$ is the actual membrane potential at time-step $t-1$ and $\mathbf{X}^t$ denotes the input vector at time-step $t$, e.g., the activation vector from the previous layer in MLPs. The output spike vector $\mathbf{S}^{t}$ is given by the Heaviside step function $\mathbb{H}(\cdot)$ in Eq. \ref{eq:lif 2}, indicating that a spike is fired when the membrane potential exceeds the potential threshold $V_{th}$. Dependent on whether the spike is fired at time-step $t$, the membrane potential $\mathbf{V}^{t}$ is set to $\mathbf{U}^{t}$ or the reset potential $V_{reset}$ through Eq. \ref{eq:lif 3}.

Since the Heaviside step function $\mathbb{H}(\cdot)$ is not differentiable, the surrogate gradient method is utilized to solve this issue, which is defined as :
\begin{equation}
    \frac{d\mathbb{H}(x)}{d x} = \frac{1}{1+\exp(-\alpha x)},
    \label{eq:lif surrogate gradient}
\end{equation}
where $\alpha$ is a predefined hyperparameter. Thus, spiking neural networks can be optimized end-to-end.

\section{Methodology}
\label{sec:methodology}

\subsection{Data encoding}
\label{sec:encoding}

\begin{figure}[t]
  \centering
  \includegraphics[width= 6cm]{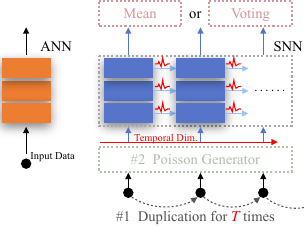}
  \caption{Conventional data encoding schemes. For direct-encoding, only the operation \#1 is necessary that it duplicates the input data $T$ times to fit the length of the temporal dimension. For Poisson-encoding, both operation \#1 and \#2 are utilized to generate the input spike train. The “Mean” or “Voting” operation is able to decode the SNN output.}
  \label{fig:data encoding}
\end{figure}

In this subsection, we explore two naive data encoding approaches for converting the input data to SNN-tailored formats, i.e., direct-encoding and Poisson-encoding. Both of them are proven to perform well in the direct learning of SNNs \cite{fang2021incorporating,han2020rmp,shrestha2018slayer,cheng2020lisnn}.

As described in Preliminaries, spiking neurons receive data with an additional dimension called the temporal dimension, which is indexed by the time-step $t$ in Eq. (\ref{eq:lif 1}-\ref{eq:lif 3}). Consequently, original ANN-formatted data need to be encoded to fill the temporal dimension as illustrated in Fig. \ref{fig:data encoding}. 
In the direct-encoding scheme, the original data is duplicated $T$ times to fill the temporal dimension, where $T$ represents the total length of the temporal dimension. As for the Poisson-encoding scheme, besides the duplication operation, it  perceives the input value as the probability and generates a spike according to the probability at each time step. 
Additionally, a decoding method is entailed for the subsequent operations of rendering, and the mean  \cite{li2021differentiable} and the voting \cite{fang2021incorporating} decoding operations are commonly considered. We employ the former approach since the latter one is designed for classification tasks \cite{diehl2015unsupervised,wu2019direct}.

Thus, with the above two encoding methods, we build two naive versions of SpikingNeRF and are able to conduct experiments on various datasets to verify the feasibility.

\subsection{Time-ray alignment (TRA)}
\label{sec:alignmen}

\begin{figure*}[t]
  \centering
  \includegraphics[width= 0.8\textwidth]{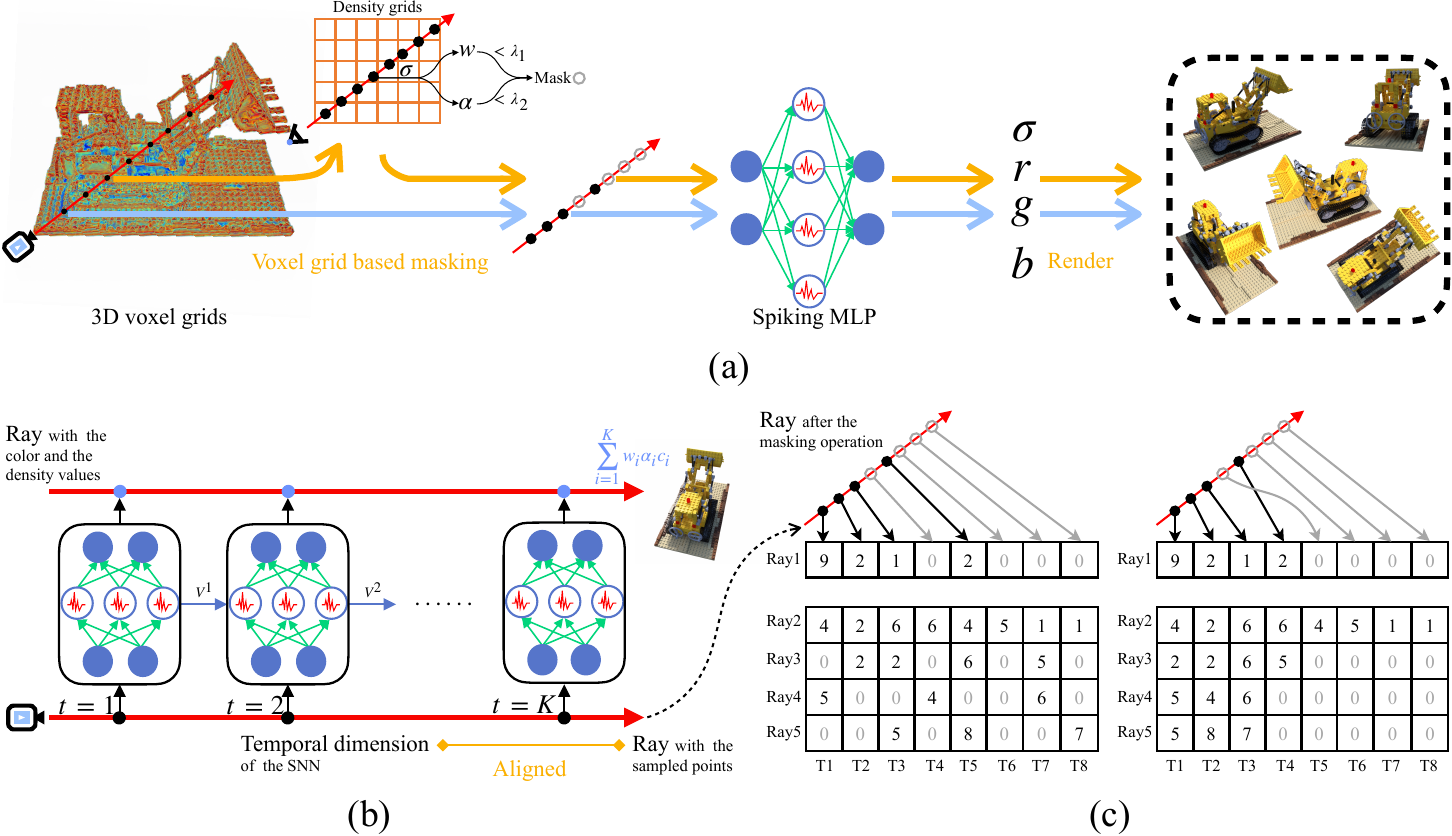}
  \caption{Overview of the proposed SpikingNeRF. (a) The rendering process of SpikingNeRF. The whole 3D volumetric parameters are stored in the voxel grids.
  The irrelevant or unimportant samples are masked before the sMLP querying. The expected scenes are rendered with the volumetric information yielded by the sMLP.
  (b) Alignment between the temporal dimension and the ray. The sMLP queries each sampled point step-by-step to yield the volumetric information. (c) Proposed temporal padding (left) and temporal condensing-and-padding (right) methods. For simplification, the channel length of the volumetric parameters is set to 1.}
  \label{fig:overview}
\end{figure*}
This subsection further explores the potential of accommodating the SNN to the NeRF rendering process in a more natural and novel way, where we attempt to retain the real-valued input data as direct-encoding does, but do not fill the temporal dimension with the duplication-based approach.

We first consider the MLP querying process in the ANN philosophy. For an expected scene to reconstruct, the volumetric parameters of sampled points, e.g., $e$ and $v$ in Eq. \ref{eq:nerf mlp2}, are packed as the input data with the shape of $[batch, c_e]$ or $[batch, c_v]$, where $batch$ represents the sample index and $c$ is the channel index of the volumetric parameters. Thus, the MLP can query these data and output the corresponding color information in parallel. However, from the geometric view, the input data should be packed as $[ray, pts, c]$, where $ray$ is the ray index and the $pts$ is the index of the sampled points. 

Obviously, the ANN-based MLP querying process can not reflect such geometric relations between the ray and the sampled points. Then, we consider the computation modality of SNNs. As illustrated in Eq. (\ref{eq:lif 1}-\ref{eq:lif 3}), SNNs naturally entail the temporal dimension to process the sequential signals. This means a spiking MLP naturally accepts the input data with the shape of $[batch, time, c]$, where $time$ is the temporal index. Therefore, we can reshape the volumetric parameters back to $[ray, pts, c]$, and intuitively match each sample along the ray to the corresponding time step:
\begin{equation}
\begin{split}
    \text{Input}_{MLP}&:=[batch, c] \\
    &\Rightarrow [ray, pts, c] \\
    &\Rightarrow [batch, time, c]:= \text{Input}_{sMLP},
\end{split}
    \label{eq:alignment}
\end{equation}
which is also illustrated in \ref{fig:overview}(b). Such an alignment
does not require any input data pre-process such as duplication \cite{zhou2022spikformer} or Poisson generation \cite{garg2021dct} as prior arts commonly do.

\subsection{Temporal condensing-and-padding (TCP)}
\label{sec:tp and tap}
The masking operation on sampled points, as illustrated in Preliminaries, makes the time-ray alignment intractable. Although such masking operation improves the rendering speed and quality by curtailing the computation cost of redundant samples, it also causes the number of queried samples on different rays to be irregular, which indicates the reshape operation of Eq. \ref{eq:alignment}, i.e., shaping into a tensor, is unfeasible on GPUs after the masking operation.

To ensure computation parallelism on GPUs, we first propose to retain the indices of those masked samples but discard their values. As illustrated in Fig. \ref{fig:overview}(c) Left, we arrange both unmasked and masked samples sequentially to the corresponding $ray$-indexed vector, and pad zeros to the vacant tensor elements. Such that, a regular-shaped input tensor is built, \emph{making GPU training feasible}. We refer to this simple approach as the temporal padding (\textbf{TP}) method.

However, TP does not handle those masked samples effectively because those padded zeros will still get involved in the following computation and cause the membrane potential of sMLP to decay, implicitly affecting the outcomes of the unmasked samples in the posterior segment of the ray. Even for a sophisticated hardware accelerator that can skip those zeros, the sparse data structure still causes computation inefficiency such as imbalanced workload \cite{zhang2020sparch}. To solve this issue, we design the temporal condensing-and-padding (TCP) scheme, which is illustrated in Fig. \ref{fig:overview}(c) Right. Different from TP, TCP completely discards the parameters and indices of the masked samples, and adjacently rearranges the unmasked sampled points to the corresponding $ray$ vector. For the vacant tensor elements, zeros are filled as TP does. Consequently, valid data is condensed to the left side of the tensor. Notably, the $ray$ dimension can be sorted according to the valid data number to further increase the density. As a result, TCP has fully eliminated the impact of the masked samples and made SpikingNeRF more hardware-friendly. 

Although such data condensing operation can incur extra overhead on hardware, \emph{a regular and condensed data structure commonly brings far more benefits to efficiency, covering the cost} \cite{zhang2020sparch}.
Such benefits not only cater for DNN hardware accelerators and GPUs, but also apply to neuromorphic hardware, as proposed and proved in PTB \cite{lee2022parallel} and STELLAR \cite{mao2024stellar}.
Therefore, we choose \emph{TCP as our mainly proposed method}.


\subsection{Overall algorithm}
\label{sbsec:overall algorithm}
This section summarizes the overall algorithm of SpikingNeRF based on the DVGO \cite{sun2022direct} framework. And, the pseudo code is given in Algorithm \ref{algo:overall algo}.

\begin{algorithm}[t]
	\renewcommand{\algorithmicrequire}{\textbf{Input:}}
	\renewcommand{\algorithmicensure}{\textbf{Output:}}
    \caption{Overall algorithm of the DVGO-based SpikingNeRF (SpikingNeRF-D) in the rendering process.}
    \label{algo:overall algo}
    \begin{algorithmic}[1]
		\REQUIRE The density and the feature voxel grids $V_\sigma$ and $V_f$, the spiking MLP $sMLP(\cdot)$,  the view direction of the camera $v$, the rays from the camera origin to the directions of $N$ pixels of the expected scene $R_{\{N\}}=\{r_1,r_2,...,r_N\}$, the number of the sampled points per ray $M$, the ground-truth RGB $\mathbf{C}=\{C_1,C_2,...,C_N\}$ of the expected $N$ pixels . 
  
		\ENSURE The expected RGB $\mathbf{\hat{C}}=\{\hat{C}_1,\hat{C}_2,...,\hat{C}_N\}$ of the expected $N$ pixels, the training loss $\mathcal{L}$.
  
        \STATE The coordinates of sampled points  $P_{\{N\times M\}}=\{p_{1,1},p_{1,2},...,p_{N,M}\}\gets Sample(R)$.

        \STATE $\alpha_{\{N \times M\}}, w_{\{N \times M\}} \gets Weigh(P, V_\sigma)$ as in Eq. \ref{eq:voxel optimization sigma} and Eq. \ref{eq:nerf weights}.
        
        \STATE Filtered coordinates $P' \gets Mask(P, \alpha, w)$ as in Eq. \ref{eq:voxel mask1}. 

        \STATE Input$_{MLP}$ $\gets ExtractFeatures(P', V_f, v)$ as described in Eq. \ref{eq:voxel optimization embed} and Eq. \ref{eq:nerf mlp2}.

        \STATE The temporal length $T \gets$ The maximum point number among the batched rays. 

        \STATE Input$_{sMLP}$ $\gets$ The \textbf{TP or TCP} transformation on Input$_{MLP}$ as described in Eq. \ref{eq:alignment} and Sec. TCP.

        \STATE The RGB values $c_{\{N,T\}} \gets \textbf{sMLP}(\text{Input}_{sMLP})$

        \STATE $\mathbf{\hat{C}} \gets Accumulate(P',\alpha, w, c)$ as in Eq. \ref{eq:voxel mask2}.

        \STATE $\mathcal{L} \gets MSE(\mathbf{C}, \mathbf{\hat{C}})$ as in Eq. \ref{eq:nerf mse}
        
    \end{algorithmic}
    Note: Dependent on the specific NeRF framework, the functions, e.g., $Sample(\cdot)$, $Mask(\cdot)$, may be different.
\end{algorithm}

As illustrated in Fig. \ref{fig:overview}(a), SpikingNeRF first establishes the voxel grids filled with learnable volumetric parameters. In the case of the DVGO implementation, two groups of voxel grids are built as the input of Algorithm \ref{algo:overall algo}, which are the density and the feature voxel grids. Given an expected scene with $N$ pixels to render, Step 1 is to sample $M$ points along each ray shot from the camera origin to the direction of each pixel. With the $N\times M$ sampled points, Step 2 queries the density grids to compute the weight coefficients, and Step 3 uses these coefficients to mask out those irrelevant points. Then, Step 4 queries the feature grids for the filtered points and returns each point's volumetric parameters. Step 5 and 6 prepare the volumetric parameters into a receivable data format for $sMLP$ with TP or TCP. Step 7 and 8 compute the RGB values for the expected scene. If a backward process is required, Step 9 calculates the MSE loss between the expected and the ground-truth scenes. 

Notably, the proposed methods, \textbf{functioning in a plug-in way}, can be applied to other NeRF frameworks, e.g., the SoTA TensoRF \cite{chen2022tensorf}, and is also orthogonal and applicable to those efficient NeRFs such as FastNeRF\cite{garbin2021fastnerf} and KiloNeRF\cite{reiser2021kilonerf}.

\section{Experiments}
\label{sec:experiments}

In this section,  we demonstrate the effectiveness of our proposed SpikingNeRF. \textbf{A)} We first build SpikingNeRF on the voxel-grid based DVGO framework \cite{sun2022direct}, and compare the proposed TRA encoding with the naive data encodings. \textbf{B)} We extend SpikingNeRF to the TensoRF framework \cite{chen2022tensorf} to show the flexibility of our method, and compare SpikingNeRF with the original DVGO and TensoRF along with other NeRF-based works in both rendering quality and energy cost. \textbf{C)} We evaluate SpikingNeRF on the \emph{neuromorphic accelerator SpikeSim} \cite{10122627} and SATA \cite{yin2022sata} and GPU to compare the hardware friendliness of the proposed TCP over TP, meanwhile showcase the energy advantage on neuromorphic accelerators over ANNs. \textbf{D)} we further discuss the alignment direction issue and extensively compare the merits of our edge-friendly spiking approach with the classic quantization method.
Note that, for text saving, we defer the results of unbounded inward-facing and forward-facing datasets and all visualizations to the appendix, and the results of SATA evaluation is also deferred along with the detailed hardware descriptions. 
For clarity, we term the DVGO-based SpikingNeRF as 
\textbf{SpikingNeRF-D} and the TensoRF-based as \textbf{SpikingNeRF-T}. If not stated otherwise, TCP is utilized by default.

\subsection{Experimental settings}
\label{sbsec:experiments}
We conduct experiments mainly on the four inward-facing datasets, including Synthetic-NeRF\cite{mildenhall2021nerf}, Synthetic-NSVF\cite{liu2020neural}, BlendedMVS\cite{yao2020blendedmvs}, and Tanks\&Temples\cite{knapitsch2017tanks}. We refer to \emph{DVGO as the ANN counterpart to SpikingNeRF-D}.
We refer to \emph{TensoRF as the ANN counterpart to SpikingNeRF-T}.
In terms of the energy computation, we follow the prior arts \cite{zhou2022spikformer, horowitz20141} to estimate the theoretical rendering energy cost in most of our experiments except for those in Tab. \ref{exp:eval on spikesim} whose results are produced by the neuromorphic accelerator SpikeSim. 
\textbf{Very detailed implementation hyper-parameters, the experiment fairness declaration, and thorough SpikeSim evaluation details are all specified in the appendix.}

\subsection{Comparisons and ablations}
\label{sbsec:compare}

\begin{table*}[t]
\centering
\caption{Comparisons with direct-encoding under different time step settings.}
\label{exp:compare with duplication-based encoding1}
\begin{adjustbox}{max width=0.88\textwidth}
\begin{threeparttable}
\begin{tabular}{l|ccc|ccc|ccc} 
\hline
Metric                           & PSNR$\uparrow$ & SSIM$\uparrow$ & 
Energy(mJ) $\downarrow$

& PSNR$\uparrow$ & SSIM$\uparrow$ & 

Energy(mJ) $\downarrow$

& PSNR$\uparrow$ & SSIM$\uparrow$ & Energy(mJ) $\downarrow$

\\ 
\hline \hline
\textit{Direct-Encoding}         & \multicolumn{3}{c|}{TimeStep=1}                                                                     & \multicolumn{3}{c|}{TimeStep=2}                                                                     & \multicolumn{3}{c}{TimeStep=4}                                                                       \\ 
\hline
Synthetic-NeRF                   & 31.22          & 0.947          & 113.03                                                            & 31.51          & 0.951          & 212.20                                                            & 31.55          & 0.951          & 436.32                                                             \\ 
\hline
Synthetic-NSVF                   & 34.17          & 0.969          & \textbf{53.73}                                                    & 34.49          & 0.971          & 104.05                                                            & 34.56          & 0.971          & 217.86                                                             \\ 
\hline \hline
\textit{TRA} & \multicolumn{9}{c}{Dynamic Time Step}                                                                                                                                                                                                                                                                            \\ 
\hline
Synthetic-NeRF                   & \textbf{31.34} & \textbf{0.949} & \textbf{110.80}                                                   & \textbf{31.59} & \textbf{0.951} & \textbf{185.78}                                                   & \textbf{31.64} & \textbf{0.952} & \textbf{308.84}                                                    \\ 
\hline
Synthetic-NSVF                   & \textbf{34.33} & \textbf{0.970} & 56.69                                                             & \textbf{34.63} & \textbf{0.972} & \textbf{98.39}                                                    & \textbf{34.57} & \textbf{0.972} & \textbf{165.17}                                                    \\
\hline
\end{tabular}
\begin{tablenotes}
      \footnotesize
    \item \emph{TRA} denotes the proposed time-ray alignment with TCP. 
\end{tablenotes}
\end{threeparttable}
 \end{adjustbox}
\end{table*}

\begin{table*}[t]
\centering
\caption{Comparisons with direct-encoding on the same sampling density levels.}
\label{exp:compare with duplication-based encoding2}
\begin{adjustbox}{max width=0.88\textwidth}
\begin{threeparttable}
\begin{tabular}{l|ccc|ccc|ccc} 
\hline
Density Level  & \multicolumn{3}{c|}{1 (Base)}                                                                       & \multicolumn{3}{c|}{2}                                                                              & \multicolumn{3}{c}{4}                                                                                \\ 
\hline
Metric                           & PSNR$\uparrow$ & SSIM$\uparrow$ & 
Energy(mJ) $\downarrow$

& PSNR$\uparrow$ & SSIM$\uparrow$ & 

Energy(mJ) $\downarrow$

& PSNR$\uparrow$ & SSIM$\uparrow$ & Energy(mJ) $\downarrow$

\\ 
\hline \hline
\multicolumn{10}{c}{\textit{Direct-Encoding}}                                                                                                                                                                                                                                                                                     \\ 
\hline
Synthetic-NeRF & 31.22          & 0.947          & 113.03                                                            & 31.40          & 0.949          & 192.81                                                            & 31.46          & 0.950          & 337.98                                                             \\ 
\hline
Synthetic-NSVF & 34.17          & 0.969          & \textbf{53.73}                                                    & 34.45          & 0.970          & \textbf{94.58}                                                    & 34.56          & 0.971          & 168.10                                                             \\ 
\hline\hline
\multicolumn{10}{c}{\textit{Time-ray Alignment with TCP}}                                                                                                                                                                                                                                                                         \\ 
\hline
Synthetic-NeRF & \textbf{31.34} & \textbf{0.949} & \textbf{110.80}                                                   & \textbf{31.59} & \textbf{0.951} & \textbf{185.78}                                                   & \textbf{31.64} & \textbf{0.952} & \textbf{308.84}                                                    \\ 
\hline
Synthetic-NSVF & \textbf{34.33} & \textbf{0.970} & 56.69                                                             & \textbf{34.63} & \textbf{0.972} & 98.39                                                             & \textbf{34.57} & \textbf{0.972} & \textbf{165.17}                                                    \\
\hline
\end{tabular}
\end{threeparttable}
 \end{adjustbox}
\end{table*}

\begin{table*}[t]
\centering
\caption{Comparisons with the ANN counterpart and other NeRF-based methods.}
\label{exp:comparison with ann baseline}
\begin{adjustbox}{max width=\textwidth}
\begin{threeparttable}
\begin{tabular}{l|cccc|cccc|cccc|cccc} 
\hline
Dataset       & \multicolumn{4}{c|}{Synthetic-NeRF}                                                                                         & \multicolumn{4}{c|}{Synthetic-NSVF}                                                                                         & \multicolumn{4}{c|}{BlendedMVS}                                                                                             & \multicolumn{4}{c}{TanksTemples}                                                                                             \\
Metric        & PSNR$\uparrow$ & SSIM$\uparrow$ & \begin{tabular}[c]{@{}c@{}}Energy$\downarrow$\\ (mJ)\end{tabular} & P/E$\uparrow$ & PSNR$\uparrow$ & SSIM$\uparrow$ & \begin{tabular}[c]{@{}c@{}}Energy$\downarrow$\\ (mJ)\end{tabular} & P/E$\uparrow$ & PSNR$\uparrow$ & SSIM$\uparrow$ & \begin{tabular}[c]{@{}c@{}}Energy$\downarrow$\\ (mJ)\end{tabular} & P/E$\uparrow$ & PSNR$\uparrow$ & SSIM$\uparrow$ & \begin{tabular}[c]{@{}c@{}}Energy$\downarrow$\\ (mJ)\end{tabular} & P/E$\uparrow$  \\ 
\hline
NeRF         & 31.01          & 0.947          & 4.5e5                                                             & 6.9e-5                & 30.81          & 0.952          & 4.5e5                                                             & 6.8e-5                & 24.15          & 0.828          & 3.1e5                                                             & 7.8e-5                & 25.78          & 0.864          & 1.4e6                                                             & 1.8e-5                 \\
Mip-NeRF      & 33.09          & 0.961          & 4.5e5                                                             & 7.4e-5                & -              & -              & -                                                                 & -                     & -              & -              & -                                                                 &           -            & -              & -              & -                                                                 &         -               \\
JaxNeRF       & 31.65          & 0.952          & 4.5e5                                                             & 7.0e-5                & -              & -              & -                                                                 & -                     & -              & -              & -                                                                 &             -          & 27.94          & 0.904          & 1.4e6                                                             & 2.0e-5                 \\
NSVF         & 31.74          & 0.953          & 16427                                                             & 1.9e-3                & 35.13          & 0.979          & 8864                                                              & 4.0e-3                & 26.90          & 0.898          & 15149                                                             & 1.8e-3                & 28.40          & 0.900          & 101443                                                            & 2.8e-4                 \\
DIVeR         & 32.32          & 0.960          & 343.96                                                            & 0.094                 & -              & -              & -                                                                 & -                     & 27.25          & 0.910          & 548.65                                                            & 0.050                 & 28.18          & 0.912          & 1930.67                                                           & 0.015                  \\

KiloNeRF       & 31.00          & 0.95          & 185.12    & 0.167      
                & 33.37          & 0.97          & 99.89      & 0.334   

                  &27.39   & 0.92         &170.71   & 0.160
                 & 28.41          & 0.91          & 723.79       & 0.039         \\

DVGO*        & 31.98          & 0.957          & 374.72                                                            & 0.085                 & 35.12          & 0.976          & 187.85                                                            & 0.187                 & \textbf{28.15} & \textbf{0.922} & 320.66                                                            & 0.088                 & 28.42          & 0.912          & 2147.86                                                           & 0.012                  \\
TensoRF*     & \textbf{33.14} & \textbf{0.963} & 641.17                                                            & 0.052                 & \textbf{36.74} & \textbf{0.982} & 465.09                                                            & 0.079                 & -              & -              & -                                                                 &             -          & \textbf{28.50} & \textbf{0.920} & 2790.03                                                           & 0.010                  \\ 
\hline

SpikingNeRF-D w/ TP                       & 31.34          & 0.949          & 111.59                                                            & 0.281                               & 34.34          & 0.970          & 57.57                                                             & 0.596                               & 27.80          & 0.912          & 97.38                                                             & 0.285                               & 28.00          & 0.892          & \textbf{483.48}                                                       & \textbf{0.057}      \\

SpikingNeRF-D w/ TCP & 31.34          & 0.949          & \textbf{110.80}                                                   & \textbf{0.283}        & 34.34          & 0.970          & \textbf{56.69}                                                    & \textbf{0.606}        & 27.80          & 0.912          & \textbf{96.37}                                                    & \textbf{0.288}        & 28.09          & 0.896          & 581.04                                                 & 0.048         \\

SpikingNeRF-T w/ TCP & 32.45          & 0.956          & 240.81    & 0.134                 & 35.76          & 0.978          & 149.98    & 0.238                 & -              & -              & -                                                                 &     -                  & 28.09          & 0.904          & 1165.90                                                           & 0.024                  \\
\hline
\end{tabular}
\begin{tablenotes}
\footnotesize
\item * denotes an ANN counterpart implemented by the official codes.
\item P/E abbreviates the “PSNR/Energy”.
\end{tablenotes}
\end{threeparttable}
\end{adjustbox}
\end{table*}

\begin{table*}[t]
        \centering
    \begin{minipage}{1\columnwidth}
        \centering
        \caption{Comparisons between TCP and TP on SpikeSim.}
        \label{exp:eval on spikesim}
        \begin{adjustbox}{max width=0.86\textwidth}
        \begin{threeparttable}
            \begin{tabular}{l|cc|cc}
            \hline
            Dataset                    & \multicolumn{2}{c|}{Synthetic-NeRF} & \multicolumn{2}{c}{Synthetic-NSVF} \\
            SpikingNeRF-D                  & w/ TCP               & w/ TP        & w/ TCP               & w/ TP       \\ \hline
            
            Latency(s)$\downarrow$     & \textbf{26.12}       & 222.22       & \textbf{13.37}       & 164.61      \\
            Energy$^+$(mJ)$\downarrow$ & \textbf{65.78}       & 559.45       & \textbf{33.68}       & 414.37      \\ \hline
            \end{tabular}
        \begin{tablenotes}
              \footnotesize
              \item $+$ denotes the energy result particularly produced by SpikeSim.
        \end{tablenotes}
        \end{threeparttable}
        \end{adjustbox}
    \end{minipage}
    \quad
    \begin{minipage}{1.\columnwidth}
        \centering
        \caption{Comparisons with temporal flip.}
        \label{exp:temporal flip}
        \begin{adjustbox}{max width=0.86\textwidth}
        \begin{threeparttable}
            \begin{tabular}{l|cc|cc}
            \hline
            Dataset                 & \multicolumn{2}{c|}{Synthetic-NeRF} & \multicolumn{2}{c}{Synthetic-NSVF} \\
            SpikingNeRF-D               & w/o TF                & w/ TF       & w/o TF               & w/ TF       \\ \hline
            PSNR$\uparrow$          & \textbf{31.34}        & 31.25       & \textbf{34.34}       & 34.15       \\
            SSIM$\uparrow$          & \textbf{0.949}        & 0.947       & \textbf{0.970}       & 0.967       \\
            Energy (mJ)$\downarrow$ & \textbf{110.80}       & 116.91      & \textbf{56.69}       & 61.08       \\ \hline
            \end{tabular}
        \begin{tablenotes}
              \footnotesize
              \item \textbf{TF} denotes temporal flip.
        \end{tablenotes}
        \end{threeparttable}
        \end{adjustbox}
    \end{minipage}
\end{table*}

\textbf{Comparisons with the conventional data encodings.} As described in Data encoding, we propose two naive versions of SpikingNeRF-D that adopt two different conventional data encoding schemes: direct-encoding and Poisson-encoding. On the one hand, Poisson-encoding, severely losing the feature information and producing \textbf{at most 24.83 PSNR} among all time-step settings and datasets, achieves far-from-acceptable synthesis quality, which indicates it does not work at all. The corresponding quantitative and qualitative results of this ineffective scheme are deferred to the appendix. On the other hand, as listed in Tab. \ref{exp:compare with duplication-based encoding1}, direct-encoding obtains good synthesis performance with only one time-step, and can achieve higher PSNR as the time step increases. Inheriting the good startup of direct-encoding, our proposed TRA shows better energy efficiency and rendering ability over direct-encoding. In Tab. \ref{exp:compare with duplication-based encoding1}, we change the TRA's time step by adjusting the default sampling density to compare with Direct-Encoding (DE) of different time steps since it is unfeasible to explicitly set TRA's time step due to its dynamic temporal length. Tab. \ref{exp:compare with duplication-based encoding1} shows that TRA has \emph{better rendering quality under the same energy levels}. We also compare TRA with  DE under the same sampling densities in Tab. \ref{exp:compare with duplication-based encoding2} for fairness, and the outcome still holds. The appendix contains the full statistics of Tab. \ref{exp:compare with duplication-based encoding1} and Tab. \ref{exp:compare with duplication-based encoding2} for each specific scene, which also show TRA consistently outperforms these conventional encodings. Conclusively,  TRA exploits SNN's temporal characteristics in 3D rendering and proves simple and effective.
\\\textbf{Quantitative comparisons with the ANN counterparts and other NeRFs.}
As shown in Tab. \ref{exp:comparison with ann baseline}, our SpikingNeRF-D with TCP can achieve a 70.79\% energy saving with a 0.53 PSNR drop on average over the ANN counterpart.
Such a trade-off between synthesis quality and energy cost is reasonable because a significant part of inference is conducted with the addition operations in the sMLP of SpikingNeRF-D rather than the multiplication operations in the original DVGO. 
On the one hand, compared with those methods, e.g., NeRF\cite{mildenhall2021nerf}, Mip-NeRF\cite{barron2021mip}, JaxNeRF\cite{deng2020jaxnerf}, that do not perform the masking operation, SpikingNeRF-D can reach orders of magnitude lower energy consumption. On the other hand, compared with those methods, e.g., NSVF\cite{liu2020neural}, DIVeR\cite{wu2022diver}, DVGO\cite{sun2022direct}, TensoRF\cite{chen2022tensorf}, that significantly exploit the masking operation,  SpikingNeRF-D can still obtain better energy efficiency and comparable synthesis quality. Even compared with KiloNeRF\cite{reiser2021kilonerf} that is aiming at fast rendering (which takes days to train), SpikingNeRF-D (that takes minutes to train) still performs better.
Furthermore, following  \cite{alyamkin20182018, lee2023mf}, we adopt the analogous PSNR/Energy to further estimate the energy efficiency of SpikingNeRF and the ANN baselines. As listed in Tab. \ref{exp:comparison with ann baseline}, SpikingNeRF-D achieves superior energy efficiency among these competitors. Given Fig. \ref{fig:performance-energy}, the superiority of our SpikingNeRF in energy efficiency is vivid.
Moreover, SpikingNeRF-T also reduces energy consumption by 62.80\% with a 0.69 PSNR drop on average. Except for Tanks\&Temples, SpikingNeRF-T outperforms DVGO in both PSNR and energy cost. Notably, SpikingNeRF-T only uses two FC layers as TensoRF does. One layer is for encoding data with full precision, the other for spiking with binary computation, leading to only half of the computation burden being tackled with the addition operations. 
This explains why SpikingNeRF-T performs slightly worse than SpikingNeRF-D in terms of energy reduction ratio.
In conclusion, these results demonstrate the effectiveness of our proposed SpikingNeRF in improving energy efficiency. 
\\\textbf{Qualitative comparisons.} Due to text limitation, we defer piles of visualizations of SpikingNeRF-D rendering results to the appendix. Basically, SpikingNeRF-D shares the analogous issues with the ANN counterpart on texture distortion. 
\\\textbf{Advantages of temporal condensing on the hardware.} To demonstrate the advantages of the proposed temporal condensing on hardware accelerators as described in Sec. TCP, \emph{we evaluate SpikingNeRF-D with TCP and TP on SpikeSim using the SpikeFlow architecture}. 
For one thing, as listed in Tab. \ref{exp:eval on spikesim}, TCP consistently outperforms TP in both inference latency and energy overhead by a significant margin over the two datasets. Specifically, The gap between TCP and TP is about an order of magnitude in both inference speed and energy cost. 
These results indicate TCP is simple but also effective at the inference stage. The same conclusion can also be drawn from the SATA \cite{yin2022sata} evaluation as shown in the appendix, which means TCP can benefit the sparsity-aware (event-driven) hardware as well. For the other, comparing the results on SpikeSim (65nm technology, Tab. \ref{exp:eval on spikesim}) with those on 45nm technology general hardware (Tab. \ref{exp:comparison with ann baseline}) further demonstrates that the proposed SpikingNeRF-D can substantially benefit from its neuromorphic computing nature on the neuromorphic hardware, achieving higher energy-efficiency over ANN baselines. Additionally, the temporal condensing will not harm the rendering quality at all as shown in Tab. \ref{exp:comparison with ann baseline}.
\\In the SpikeSim evaluation, \emph{the temporal condensing operation is done off-chip}. This causes on-chip computation can fully benefit from the dense data, thus accounting for the huge performance gap between TCP and TP. But note that due to the pipeline mechanism, the latency of such off-chip operation can be easily covered. To evaluate the substantial overhead and merits that temporal condensing can actually bring, we showcase the training and inference time on Synthetic-NeRF on single A100 GPU.  
\begin{figure}[h]
  \centering
  \includegraphics[width= 8cm]{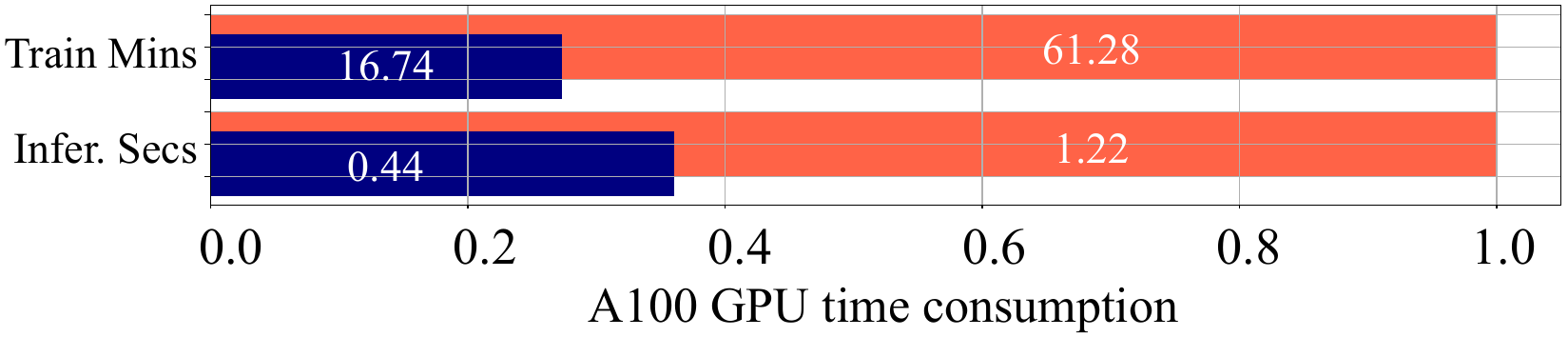}
\end{figure}
\\In the above figure, the orange bar is the time consumption of TP, while the blue one is that of TCP. Even roughly realizing the temporal condensing with PyTorch can still bring significant time-saving at the rendering stage on GPU.
\begin{table}
    \centering
    \caption{Comparisons with Quantized NeRF (qNeRF).}
    \label{exp:compare with lsq}
        \begin{adjustbox}{max width=0.36\textwidth}
        \begin{threeparttable}
        \begin{tabular}{c|cc|cc}
        \hline
        Dataset & \multicolumn{2}{c|}{Synthetic-NeRF}       & \multicolumn{2}{c}{Synthetic-NSVF}        \\ \hline
        Metric  & PSNR$\uparrow$          & Energy$\downarrow$         & PSNR$\uparrow$            & Energy$\downarrow$          \\ \hline
        qNeRF   & 31.24          & 167.67         & 34.13          & 78.54          \\ \hline
        ours   & \textbf{31.34} & \textbf{110.80} & \textbf{34.33} & \textbf{56.69} \\ \hline
        \end{tabular}
        \end{threeparttable}
        \end{adjustbox}
\end{table}
\\\textbf{Discussion of the alignment direction.} 
The radiance accumulation originally has no direction but SNN has, so it's necessary to discuss the time-ray alignment direction.
We propose temporal flip to empirically decide the alignment direction since the querying direction of sMLP along the camera ray will affect the inference outcome. Tab. \ref{exp:temporal flip} lists the experimental results of SpikingNeRF-D with and without temporal flip, i.e., with the consistent and the opposite directions. Distinctly, keeping the direction of the temporal dimension consistent with that of the camera ray outperforms the opposite case on the two datasets in synthesis performance and energy efficiency. Therefore, the consistent alignment direction is important in SpikingNeRF.
\\\textbf{Extensive comparisons with quantized ANN baselines.} To further demonstrate the merits of SpikingNeRF compared to the quantized NeRF version, 
we quantize the activation of DVGO to spike-bit, i.e., 1-bit, with the renowned LSQ \cite{esser2020learned}, and compare it with SpikingNeRF-D in Tab. \ref{exp:compare with lsq}. The results show that SpikingNeRF-D outperforms the quantized ANN version on the two datasets in both synthesis quality and energy consumption, indicating SNNs do have an advantage over ANNs in scenario of ultra-low-energy computation.

\section{Conclusion}
\label{sec:conclusion}
This paper proposes SpikingNeRF that accommodates the spiking neural network to reconstructing real 3D scenes for the first time, improving energy efficiency. TRA is developed to encode sampled points, seamlessly combining the temporal characteristic of SNNs with the radiance ray. TCP is further proposed to improve hardware friendliness. Thorough experiments are conducted to prove the effectiveness.

\bibliography{aaai25}

\end{document}